\documentclass[journal]{IEEEtai}

\usepackage[colorlinks,urlcolor=blue,linkcolor=blue,citecolor=blue]{hyperref}

\usepackage{color,array}

\usepackage{amssymb}
\usepackage{amsmath}
\usepackage{amsthm}
\usepackage{dsfont}
\usepackage{bbm}
\usepackage{booktabs} % For formal tables
\usepackage{chemfig}
\usepackage{graphicx}
\usepackage{cite}
\usepackage{mathtools}
\usepackage[normalem]{ulem}

\usepackage{url}
\usepackage{multirow}
\usepackage{float}
\usepackage[ruled, vlined,algo2e,linesnumbered]{algorithm2e}
\usepackage{tikz}
\usepackage{xcolor}
\usepackage{tkz-euclide,subfigure}
\usetikzlibrary{shapes}
\newlength{\R}
\usepackage{pgfplots}
\pgfplotsset{compat=newest}
\usetikzlibrary{shapes.arrows}
\usetikzlibrary{arrows.meta}
\usetikzlibrary{trees,arrows}

\makeatletter
\newcommand{\nosemic}{\renewcommand{\@endalgocfline}{\relax}}% Drop semi-colon ;
\newcommand{\dosemic}{\renewcommand{\@endalgocfline}{\algocf@endline}}% Reinstate semi-colon ;
\newcommand{\comment}[1]{}

\newcommand\footnoteref[1]{\protected@xdef\@thefnmark{\ref{#1}}\@footnotemark}

\newcommand{\RN}[1]{%
	\textup{\uppercase\expandafter{\romannumeral#1}}%
}

\DeclareMathAlphabet{\mathcal}{OMS}{cmsy}{m}{n}

\newtheorem{definition}{Definition}

\begin{document}

	\title{Multi-scale Wasserstein Shortest-path Graph Kernels for Graph Classification}
	
	\sloppy
	
	\author{Wei~Ye, Hao~Tian, and Qijun~Chen,~\IEEEmembership{Senior Member,~IEEE}
		\thanks{This work was supported partially by the National Natural Science Foundation of China (grant \# 62176184), the National Key Research and Development Program of China (grant \# 2020AAA0108100), and the Fundamental Research Funds for the Central Universities of China.}
		\thanks{The authors are with the College of Electronic and Information Engineering, Tongji University, Shanghai 201804 China (e-mail: \{yew, 2133036, qjchen\}@tongji.edu.cn).} 
		\thanks{This paragraph will include the Associate Editor who handled your paper.}
	}

	\markboth{Journal of IEEE Transactions on Artificial Intelligence, Vol. 00, No. 0, Month 2020}
	{Wei Ye \MakeLowercase{\textit{et al.}}: Multi-scale Wasserstein Shortest-path Graph Kernels for Graph Classification}

	\maketitle
	
	\begin{abstract}
		Graph kernels are conventional methods for computing graph similarities. However, the existing R-convolution graph kernels cannot resolve both of the two challenges: 1) Comparing graphs at multiple different scales, and 2) Considering the distributions of substructures when computing the kernel matrix. These two challenges limit their performances. To mitigate both of the two challenges, we propose a novel graph kernel called the Multi-scale Wasserstein Shortest-Path graph kernel (MWSP), at the heart of which is the multi-scale shortest-path node feature map, of which each element denotes the number of occurrences of the shortest path around a node. The shortest path is represented by the concatenation of all the labels of nodes in it. Since the shortest-path node feature map can only compare graphs at local scales, we incorporate into it the multiple different scales of the graph structure, which are captured by the truncated BFS trees of different depths rooted at each node in a graph. We use the Wasserstein distance to compute the similarity between the multi-scale shortest-path node feature maps of two graphs, considering the distributions of shortest paths. We empirically validate MWSP on various benchmark graph datasets and demonstrate that it achieves state-of-the-art performance on most datasets.
	\end{abstract}

   \begin{IEEEImpStatement}
   	Many real-world graphs exhibit the small-world property, which means the average shortest-path length is small. The shortest-path graph kernel (SP) is suitable for computing the similarities between such graphs. However, it represents the shortest path as a triplet that contains the labels of the source and sink nodes and the length of this shortest path as components. This simple representation does not take the labels of the other constituent nodes in the shortest path into account and thus loses information. In addition, SP compares graphs without considering the multiple different scales of the graph structure which is common in real-world graphs. Moreover, the sum aggregation of the shortest path similarities in SP ignores valuable information such as the distribution of the shortest paths. The Multi-scale Wasserstein Shortest-Path graph kernel (MWSP) introduced in this paper overcomes these challenges and generates improved results on various graph benchmark datasets, making it quite competitive on the graph classification task.
    \end{IEEEImpStatement}


%\sloppy

\section{Introduction}\label{intro}

 \IEEEPARstart{G}{raph-structured} data is ubiquitous in many scenarios, such as protein-protein interaction networks, social networks, citation networks, and cyber-physical systems. One interesting and fundamental problem with graph-structured data is quantifying their similarities, which can be measured by graph kernels. Most of the graph kernels belong to the family of the R-convolution kernels~\cite{haussler1999convolution} that decompose graphs into substructures and compare them. There are many kinds of substructures used in graph kernels, such as walks~\cite{gartner2003graph,vishwanathan2010graph,zhang2018retgk}, paths~\cite{borgwardt2005shortest,ye2019tree++}, subgraphs~\cite{shervashidze2009efficient,costa2010fast,horvath2004cyclic,kondor2016multiscale}, and subtree patterns~\cite{ramon2003expressivity,mahe2009graph,shervashidze2009fast,shervashidze2011weisfeiler,da2012tree,togninalli2020wasserstein}.

\begin{figure}[!htp]
	\centering
	\includegraphics[width=0.45\textwidth]{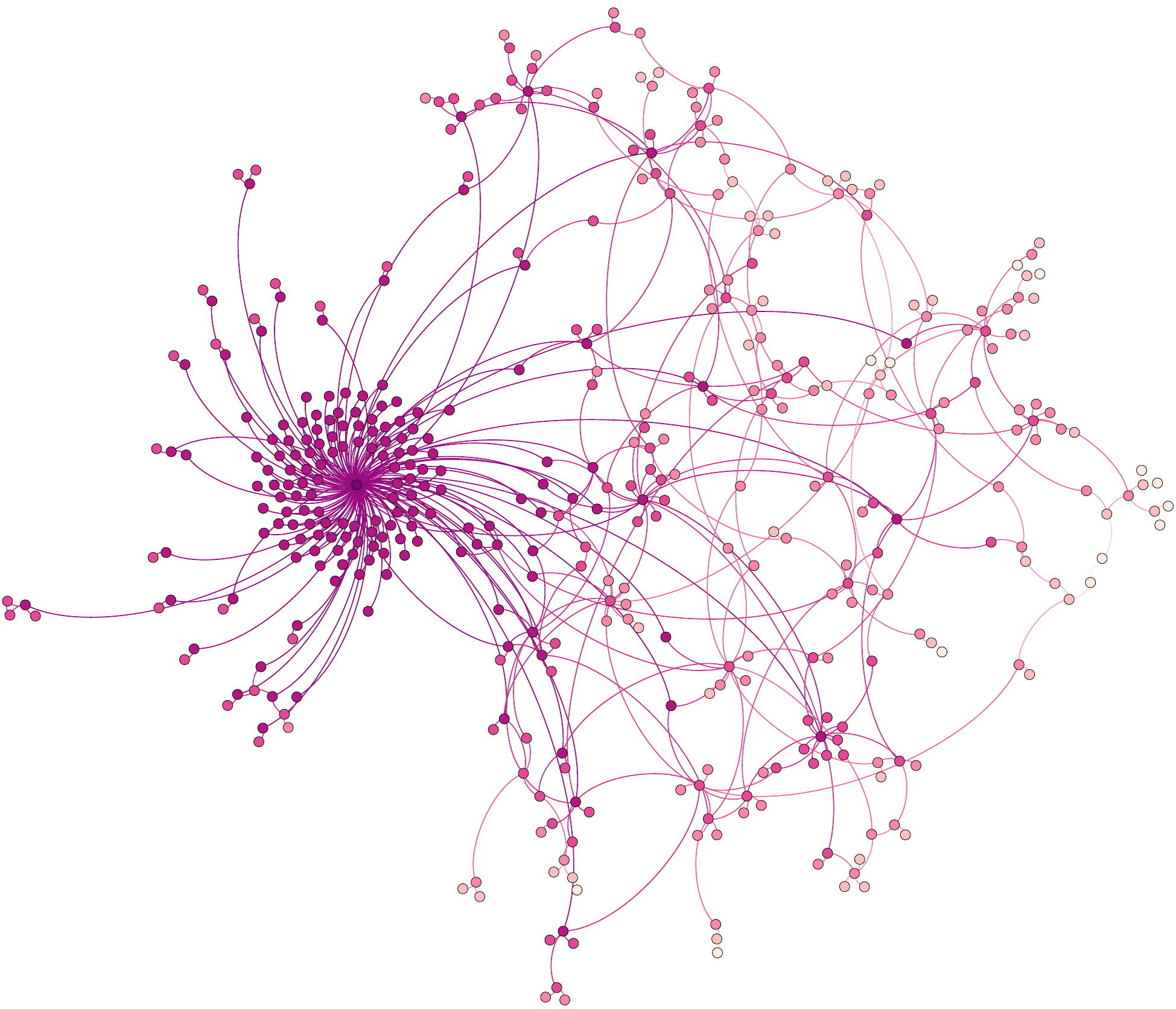}
	\caption{A graph of a post on Reddit. Each node represents a user. Two users are connected by an edge if at least one of them responds to the other's comment. The image is drawn with Gephi.}
	\label{fig:reddit}
\end{figure}

The existing R-convolution graph kernels cannot resolve both of the two challenges. Firstly, when computing the similarity between two graphs, many of them count the number of occurrences of each unique substructure in each graph and aggregate the occurrences. Simple aggregation operation such as sum aggregation neglects the distributions of substructures in each graph, which might restrict their ability to capture complex relationships between substructures. Secondly, many of them cannot compare graphs at multiple different scales. For example, WL~\cite{shervashidze2009fast,shervashidze2011weisfeiler} computes graph similarity at coarse scales, because subtrees only consider the whole neighborhood structures of nodes rather than single edges between two nodes; SP~\cite{borgwardt2005shortest} computes the graph similarity at fine scales because shortest paths do not take neighborhood structures into account. In many real-world scenarios, graphs such as social networks have structures at multiple different scales. Figure~\ref{fig:reddit} shows an online discussion thread on Reddit\footnote{\url{https://www.reddit.com/r/AskReddit}}. Nodes represent users and edges represent that at least one of the connected two users responds to the other's comment. We can see that this graph contains structures at multiple different scales, such as chains, rings, and stars. Graph kernels should not only differentiate the overall shape of graphs but also their small substructures. 

Recently, some graph kernels~\cite{togninalli2020wasserstein,kondor2016multiscale} have been proposed, only solving one of these two challenges. For example, WWL~\cite{togninalli2020wasserstein} adopts the Weisfeiler-Lehman test of graph isomorphism~\cite{leman1968reduction} to relabel each node. The new label of a node corresponds to the subtree pattern around it. WWL considers node labels as categorical embeddings. The similarity between two graphs is computed by the Wasserstein distance, capturing the distributions of subtree patterns. However, WWL cannot compare graphs at fine scales because of the utilization of the subtree patterns as used in WL. MLG~\cite{kondor2016multiscale} can compute graph similarity at multiple different scales by applying recursively the Feature space Laplacian Graph (FLG) kernel on a hierarchy of nested subgraphs. FLG is a variant of the Bhattacharyya kernel~\cite{kondor2003kernel} that computes the similarity between two graphs by using their Laplacian matrices. The sizes of subgraphs increase with the increasing level of the hierarchy, including the different scales of the graph structure around a node. Nevertheless, MLG does not take the distributions of node embeddings (denoted by the rows of the graph Laplacian matrix) into account when computing graph similarity. There still exists no R-convolution graph kernels that can resolve both of these two challenges.

Considering that many real-world graphs have the small-world property~\cite{watts1998collective}, i.e., the mean shortest-path length is small, the shortest-path graph kernel (SP)~\cite{borgwardt2005shortest} fits these kinds of graphs well. We study SP in this paper. The substructure used in SP is the shortest path which is represented by a triplet whose elements are the labels of source and sink nodes of this shortest path and its length. The triplet representation is coarse and may lose information because it neglects the labels of the intermediate nodes in the shortest path. In this paper, each shortest path is denoted by the concatenation of the labels of all the nodes in this shortest path rather than the simple triplet representation. Guided by the small-world property, we build a truncated BFS (Breadth-First Search) tree of depth $d$ rooted at each node in a graph and only consider the shortest paths starting from the root and ending in any other node in the truncated BFS tree. In this way, the maximal length of the shortest path in a graph is limited to $d$. To mitigate the two challenges, we propose a multi-scale shortest-path graph/node feature map, capturing the multiple different scales of the graph structure. The shortest-path graph/node feature map is constructed by counting the number of occurrences of shortest paths in a graph or around a node. As aforementioned, SP cannot compare graphs at multiple different scales. To integrate the multiple different scales of the graph structure into the shortest-path graph/node feature map, we allow components in the shortest path to be BFS trees of different depths. To represent such kind of shortest paths, we project each BFS tree into an integer label by an injective hash method, i.e., hashing the same BFS trees to the same integer label and different BFS trees to different integer labels. Thus, the representation of such kind of shortest paths is just the concatenation of the integer labels of all BFS trees in this shortest path. We concatenate the multiple different scales of shortest-path graph/node feature maps and adopt the Wasserstein distance to compute graph similarity, capturing the distributions of shortest paths.

Our main contributions can be summarized as follows:
\begin{itemize}
    \item We build a truncated BFS tree of limited depth rooted at each node to restrict the maximal length of the shortest paths in a graph, guided by the small-world property. Each shortest path is represented by the concatenation of all its node labels rather than the simple triplet representation, considering all information (node labels) in this shortest path.
	\item We propose the multi-scale shortest-path graph/node feature map that incorporates the different scales of the graph structure into the shortest-path graph/node feature map and can compare graphs at multiple different scales. The different scales of the graph structure are captured by the truncated BFS trees of varying depths rooted at each node.
	\item We develop a graph kernel called MWSP (for Multi-scale Wasserstein Shortest-Path graph kernel) that employs the Wasserstein distance to compute the similarity between the multi-scale shortest-path node feature maps of two graphs, capturing the distributions of shortest paths. 
	\item We show in the experiments that our graph kernel MWSP achieves the best classification accuracy on most benchmark datasets. 
\end{itemize}

The rest of the paper is organized as follows: We discuss related work in the following section. Section 3 describes the preliminaries. Section 4 introduces our approach, including the shortest-path graph/node feature map, multi-scale shortest-path graph/node feature map, and our graph kernel MWSP. Section 5 compares MWSP with other graph kernels for graph classification on the benchmark datasets. Section 6 concludes the paper.

\section{Related Work}\label{relatedWork}

Graph kernels can be categorized by the substructures they use, e.g., graph kernels based on random walks and paths, graph kernels based on subgraphs, and graph kernels based on subtree patterns.

Random walk graph kernels~\cite{gartner2003graph,vishwanathan2010graph} decompose graphs into their substructures random walks and compute the number of matching random walks from two graphs, which can be transformed to perform random walks on the direct product graph of two input graphs. G\"{a}rtner et al.~\cite{gartner2003graph} introduce the concept of direct product graph and derive two closed-form random walk graph kernels, which are called the geometric kernel and the exponential kernel, respectively. The former kernel is based on the personalized PageRank diffusion~\cite{andersen2006local} while the latter kernel is based on the heat kernel diffusion~\cite{chung2007heat}. The random walk has closed-form expressions but its time complexity is at least $\mathcal{O}(|\mathcal{V}|^3)$~\cite{vishwanathan2010graph}, where $|\mathcal{V}|$ is the number of nodes. RetGK~\cite{zhang2018retgk} proposes a graph kernel that is applicable to both attributed and non-attributed graphs. Each graph is represented by a multi-dimensional tensor that is composed of the return probability vectors of random walks starting from and ending in the same nodes and other discrete or continuous attributes of nodes. The multi-dimensional tensor is embedded into the Hilbert space by maximum mean discrepancy (MMD)~\cite{gretton2012kernel} and the random walk graph kernels are computed. However, RetGK may face the halting problem~\cite{sugiyama2015halting}, i.e., the return probability of short random walks dominates that of the long random walks, which leads to that graph similarity depends on the short random walk similarity. Borgwardt et al.~\cite{borgwardt2005shortest} propose the shortest-path graph kernel (SP) that decomposes graphs into shortest paths between each pair of nodes in graphs. SP counts the number of pairs of matching shortest paths that have the same triplet representation in two graphs. Just as described in Section~\ref{intro}, the triplet representation of the shortest path used in SP may lose information and SP cannot compare graphs at multiple different scales.

The graphlet kernel (GK)~\cite{shervashidze2009efficient} decomposes graphs into subgraphs called graphlets~\cite{prvzulj2004modeling}. Nodes in graphlet may be not fully connected. The isolated nodes function as noises, which decreases GK's performance. The number of nodes a graphlet can have is restricted to five for the efficiency of enumeration. Even by this restriction, exhaustive enumeration of all graphlets in a graph is still prohibitively expensive. Thus, Shervashidze et al.~\cite{shervashidze2009efficient} propose to speed up the enumeration by using the method of random sampling. Neighborhood Subgraph Pairwise Distance Kernel (NSPDK)~\cite{costa2010fast} decomposes a graph into all pairs of neighborhood subgraphs of small radium at increasing distances. To speed up the kernel matrix computation, NSPDK uses a fast graph invariant string encoding for the pairs of neighborhood subgraphs. This graph invariant string encoding does not manage the hash collision problem and thus is not injective. Two nodes with different labels and local graph structures may be hashed to the same new label, reducing the accuracy of graph similarity. Multiscale Laplacian Graph kernel (MLG)~\cite{kondor2016multiscale} can compare graphs at multiple different scales of the graph structure. To solve the graph invariance problem, the feature space Laplacian graph kernel (FLG) is developed. Then, each node is associated with a subgraph centered around it, and the FLG kernel between each pair of these subgraphs is computed. To compute graph similarity at multiple different scales, the size of subgraphs is increased and the FLG kernel is recursively applied to these larger subgraphs. However, MLG does not consider the distributions of subgraphs when computing graph kernels.

Weisfeiler-Lehman subtree kernel (WL)~\cite{shervashidze2009fast,shervashidze2011weisfeiler} augments each node using the first-order Weisfeiler-Lehman test of graph isomorphism~\cite{leman1968reduction}. It concatenates the labels of a node and all its neighboring nodes into a string and hashes the concatenated labels into new labels, aggregating neighborhood information around a node. The node label augmentation in WL is referred to as color refinement. Colors (labels) diverge too fast, which causes graph similarity too coarse. GWL~\cite{bause2022gradual} is proposed to slow the color divergence, using node neighborhood clustering for augmentation. However, this augmentation method is not injective. GAWL~\cite{nikolentzos2023graph} first aligns nodes in different graphs if these two nodes have the same labels assigned by the first-order Weisfeiler-Lehman test of graph isomorphism. The adjacency matrix of each graph is then reordered by the permutation matrix generated from the alignment. Graph similarity is computed by applying a linear kernel on the aligned adjacency matrices. WWL~\cite{togninalli2020wasserstein} extends WL to graphs with continuous attributes by considering each graph as a set of node embeddings and using the Wasserstein distance to compute the similarity between the two sets of node embeddings. Each node is associated with two kinds of features, one of which is the subtree feature map generated by WL and the other of which is continuous attributes. The continuous attributes of each node are propagated to its neighboring nodes by an average aggregation operator, which is similar to the message-passing mechanism of graph neural networks~\cite{kipf2016semi,li2020fast,wang2020haar,huang2022graph}. OA~\cite{kriege2016valid} first generates graph hierarchies and then computes the optimal assignment kernels, which are guaranteed positive semidefinite. The optimal assignment kernels are integrated into the WL to improve its performance. FWL~\cite{schulz2021graph} is based on a filtration kernel that compares Weisfeiler-Lehman subtree feature occurrence distributions (filtration histograms) over sequences of nested subgraphs, which are generated by sequentially considering all the edges whose weights are less than a changing threshold. This strategy not only provides graph features on different levels of granularities but also considers the life span of graph features. FWL's expressiveness is more powerful than the ordinary WL. Since the subtree pattern used in these six graph kernels is not aware of single edges, all of them cannot compare graphs at fine scales.

Recently, researchers have tried to go beyond the R-convolution graph kernels. DGK utilizes word2vec~\cite{mikolov2013efficient} to learn latent representations for substructures, such as graphlets, shortest paths, and subtree patterns. The similarity matrix between substructures is computed and integrated into the kernel matrix. If the number of substructures is large, the time complexity of computing the pairwise substructure similarity is high. Nikolentzos et al.~\cite{nikolentzos2017matching} use the eigenvectors of the graph adjacency matrices as graph embeddings and propose two graph kernels to compute the similarity between two graph embeddings. The first graph kernel uses the Earth Mover's Distance~\cite{rubner2000earth} to compute the distance between two graph embeddings; and the second one uses the Pyramid Match kernel~\cite{lazebnik2006beyond,grauman2007pyramid} to find an approximate correspondence between two graph embeddings. However, eigenvectors cannot capture the multiple different scales of the graph structure. HTAK~\cite{bai2022hierarchical} transitively aligns the vertices between graphs through a family of hierarchical prototype graphs, which are generated by the k-means clustering method. The number of cluster centroids needs to be preassigned and it is hard to set properly for different graphs. Graph Neural Tangent Kernel (GNTK)~\cite{du2019gntk} is inspired by the connections between over-parameterized neural networks and kernel methods~\cite{jacot2018neural,arora2019exact}. It is equivalent to an infinitely wide GNN trained by gradient descent. GNTK is proven to learn a class of smooth functions on graphs. GraphQNTK~\cite{tanggraphqntk} is an extension of GNTK by introducing the attention mechanism and quantum computing into GNTK's structure and computation. Both GNTK and GraphQNTK are more powerful than R-convolution graph kernels because of the high representation power of neural networks. Nevertheless, they may be overparameterized and overfitting on small graphs.

\section{Preliminaries}
\subsection{Notations}\label{pre}
In this work, the row vector is denoted by boldface lowercase letter (e.g.,\ $\mathbf{x}=[x_1,\ldots, x_n]$). We consider an undirected labeled graph $\mathcal{G}=(\mathcal{V},\mathcal{E}, l)$, where $\mathcal{V}$ is a set of nodes of size $|\mathcal{V}|$, $\mathcal{E}$ is a set of graph edges of size $|\mathcal{E}|$, and $l: \mathcal{V}\rightarrow \Sigma$ is a function that assigns labels from a set of positive integers $\Sigma$ to nodes. Without loss of generality, $\lvert\Sigma\rvert\leq |\mathcal{V}|$. An edge $e$ is denoted by two nodes $u,v$ that are connected to it. The depth of a BFS tree is the maximal length of the shortest paths between the root and any other node in the BFS tree. 

Let $\mathcal{X}$ be a non-empty set and let $\kappa: \mathcal{X} \times \mathcal{X} \rightarrow \mathbb{R}$ be a function on $\mathcal{X}$. If there is a real Hilbert space $\mathcal{H}$ and a mapping $\phi:  \mathcal{X} \rightarrow \mathcal{H}$ satisfying that $\kappa(x, y) = \langle\phi(x), \phi(y)\rangle$ for all $x$, $y$ in $\mathcal{X}$, where $\langle\cdot, \cdot\rangle$ denotes the inner product of $\mathcal{H}$, then $\kappa$ is a kernel on $\mathcal{X}$. $\phi$ is called a feature map and $\mathcal{H}$ is called a feature space. $\kappa$ is a symmetric and positive-semidefinite kernel, which can be inputted into support vector machines (SVM) for classification. Similarly for graphs, let $\phi(\mathcal{G})$ be a mapping from a graph to a vector, each element of which denotes the frequency of the substructures (such as shortest path) in graph $\mathcal{G}$. Then, the kernel on two graphs $\mathcal{G}_1$ and $\mathcal{G}_2$ is defined as $\kappa(\mathcal{G}_1, \mathcal{G}_2) = \langle\phi(\mathcal{G}_1), \phi(\mathcal{G}_2)\rangle$, which denotes the sum of substructure similarities. R-convolution graph kernels use such kind of aggregation of substructure similarities, ignoring the distribution of substructures. In this paper, we adopt the Wasserstein distance to better compute the graph similarity, capturing the distribution of the substructures.

\subsection{Wasserstein Distance}

The Wasserstein distance is a distance function defined between probability distributions on a given metric space $M$. Let $\mu$ and $\nu$ be two probability distributions on $M$ and $g$ be a ground distance such as the Euclidean distance. The $p$-th ($p\geq1$) Wasserstein distance between $\mu$ and $\nu$ is defined as follows:

\begin{equation}
\label{equ:wd}
W_p(\mu,\nu):=\left(\inf_{\gamma\in\Gamma(\mu,\nu)}\int_{M\times M}g(x,y)^p\mathsf{d}\gamma(x,y)\right)^{1/p}
\end{equation}
where $\Gamma(\mu,\nu)$ denotes the collection of all probability distributions on $M\times M$ with marginal distributions $\mu$ and $\nu$, respectively.

The one-dimensional ($p=1$) Wasserstein distance is equivalent to the Earth Mover's Distance~\cite{rubner2000earth}. Intuitively, if $\mu$ and $\nu$ are viewed as a unit amount of earth piled on $M$, $W_1(\mu,\nu)$ is the minimum cost of turning one pile into the other, which is assumed to be the amount of earth moved times the distance it has to be moved. In this paper, each graph is represented as a finite set of node feature maps. Thus, we use the one-dimensional Wasserstein distance because it supports the distance computation between two sets of vectors. We use sum instead of integral in the computation of $W_1(\mu,\nu)$, which can be formulated and solved as an optimal transport problem~\cite{villani2009optimal}:

\begin{equation}
\label{equ:optimal_transport}
\begin{split}
\min&\sum_{i=1}^{|\mathcal{V}|_1}\sum_{j=1}^{|\mathcal{V}|_2}f_{i,j}g(\mathbf{x}_i,\mathbf{y}_j)\\
&\mbox{subject to}\\
&\sum_{i=1}^{|\mathcal{V}|_1}f_{i,j}=\frac{1}{|\mathcal{V}|_2}\;\;\forall j\in\left\lbrace 1,\ldots,|\mathcal{V}|_2\right\rbrace \\
&\sum_{j=1}^{|\mathcal{V}|_2}f_{i,j}=\frac{1}{|\mathcal{V}|_1}\;\;\forall i\in\left\lbrace 1,\ldots,|\mathcal{V}|_1\right\rbrace \\
&f_{i,j}\geq 0\;\;\forall i\in\left\lbrace 1,\ldots,|\mathcal{V}|_1\right\rbrace,\;\forall j\in\left\lbrace 1,\ldots,|\mathcal{V}|_2\right\rbrace
\end{split}
\end{equation}
where $|\mathcal{V}|_1$ and $|\mathcal{V}|_2$ stand for the number of nodes in graphs $\mathcal{G}_1$ and $\mathcal{G}_2$, respectively. $\mathbf{x}_i$ is the feature map of node $v_i$ in graph $\mathcal{G}_1$ and $\mathbf{y}_j$ is the feature map of node $u_j$ in graph $\mathcal{G}_2$. $f_{i,j}$ is the $(i,j)$-th element of the flow matrix $\mathbf{F}\in\mathbb{R}^{|\mathcal{V}|_1\times |\mathcal{V}|_2}$. $g(\mathbf{x}_i,\mathbf{y}_j)=\lVert\mathbf{x}_i-\mathbf{y}_j\rVert_2$ denotes the Euclidean distance between two node feature maps. 

After finding the optimal flow matrix $\mathbf{F}$ that minimizes the above objective function, $W_1(\mathbf{X},\mathbf{Y})$ between the node feature maps of graphs $\mathcal{G}_1$ and $\mathcal{G}_2$ can be computed as follows:
\begin{equation}
\label{equ:w1}
W_1(\mathbf{X},\mathbf{Y})=\frac{\sum_{i=1}^{|\mathcal{V}|_1}\sum_{j=1}^{|\mathcal{V}|_2}f_{i,j}g(\mathbf{x}_i,\mathbf{y}_j)}{\sum_{i=1}^{|\mathcal{V}|_1}\sum_{j=1}^{|\mathcal{V}|_2}f_{i,j}}
\end{equation}

\section{Multi-scale Wasserstein Shortest-path Graph Kernels}

\subsection{Shortest-Path Graph Feature Map}
The shortest-path graph kernel (SP)~\cite{borgwardt2005shortest} uses a triplet that contains the labels of the source and sink nodes and the length of the shortest path as its representation. This coarse representation loses the information of the intermediate nodes. To alleviate this situation, we consider all the information of nodes in the shortest path. We first describe the representation of the shortest path used in our graph kernel as follows:
\begin{definition}[Shortest-path Representation]
	Given an undirected and labeled graph $\mathcal{G}=(\mathcal{V},\mathcal{E}, l)$, rooted at each node $v \in\mathcal{V}$, we construct a truncated BFS tree $\mathcal{T}_v=(\mathcal{V}',\mathcal{E}', l)$ ($\mathcal{V}'\subseteq\mathcal{V}$ and $\mathcal{E}'\subseteq\mathcal{E}$) of depth $d$. For every node $v'$ in this BFS tree,  the shortest path is constructed by concatenating all the distinct nodes and edges from the root $v$ to $v'$, i.e., $P=``v,v_1,v_2,\ldots,v'"$. The shortest-path representation is the concatenated labels of all the nodes in this shortest path, i.e., $l(P)=``l(v),l(v_1),l(v_2),\ldots,l(v')"$.
\end{definition}

\begin{figure}[htb]
	\hspace*{\fill}
	\centering
	\subfigure[An undirected labeled graph $\mathcal{G}_1$.]{
		\begin{tikzpicture}
			\begin{scope}[scale=0.9,every node/.style={circle,draw}]
				\node (A) at (0,1) {$4$};
				\node (B) at (0,0) {$1$};
				\node (C) at (0,2) {$3$};
				\node (D) at (-1,2) {$2$};
				\node (E) at (1,1) {$3$};
				\node (G) at (2,1) {$1$};

			\end{scope}
		
		\begin{scope}[scale=0.9,>={Stealth[black]},
			every node/.style={fill=white,circle},
			every edge/.style={draw=black,very thick}]
			\foreach \from/\to in {A/B,A/C,A/E,D/C,G/E}
			\draw[] (\from) -- (\to);
		\end{scope}
		\end{tikzpicture}
	}
	\hfill
	\hfill
	\centering
	\subfigure[An undirected labeled graph $\mathcal{G}_2$.]{
		\begin{tikzpicture}
		\begin{scope}[scale=0.9,every node/.style={circle,draw}]
			\node (A) at (0,1) {$4$};
			\node (B) at (1,1) {$1$};
			\node (C) at (-1,1) {$3$};
			\node (D) at (-1,0) {$2$};
			\node (E) at (0,2) {$3$};
			\node (G) at (2,1) {$1$};
			
		\end{scope}
		
		\begin{scope}[scale=0.9,>={Stealth[black]},
			every node/.style={fill=white,circle},
			every edge/.style={draw=black,very thick}]
			\foreach \from/\to in {A/B,A/C,A/E,A/D,D/C,G/B}
			\draw[] (\from) -- (\to);
		\end{scope}
		\end{tikzpicture}
	}
	\hspace*{\fill}
	
	\hspace*{\fill}
	\centering
	\subfigure[A truncated BFS tree of depth one rooted at the node with label 4 in graph $\mathcal{G}_1$.]{
		\begin{tikzpicture}
		[scale=0.9,level distance=8.66mm,
		every node/.style={draw, circle, minimum size=1.5em,inner sep=1},
		level 1/.style={sibling distance=18mm},
		level 2/.style={sibling distance=18mm}]
		\node {4}
		child {node {1}
			child[missing]
		}
		child {node {3}
			child[missing]
		}
		child {node {3}
			child[missing]
		};
		\end{tikzpicture}
	}
	\hfill
	\hfill
	\centering
	\subfigure[A truncated BFS tree of depth one rooted at the node with label 4 in graph $\mathcal{G}_2$.]{
		\begin{tikzpicture}
		[scale=0.9,level distance=8.66mm,
		every node/.style={draw, circle, minimum size=1.5em,inner sep=1},
		level 1/.style={sibling distance=12mm},
		level 2/.style={sibling distance=12mm}]
		\node {4}
		child {node {1}
			child[missing]
		}
	    child {node {2}
	    	child[missing]
	    }
        child {node {3}
        	child[missing]
        }
		child {node {3}
			child[missing]
		};
		\end{tikzpicture}
	}
	\hspace*{\fill}
	\caption{Illustration of the shortest paths in graphs. $\Sigma=\{1,2,3,4\}$.}
	\label{fig:example}
\end{figure}
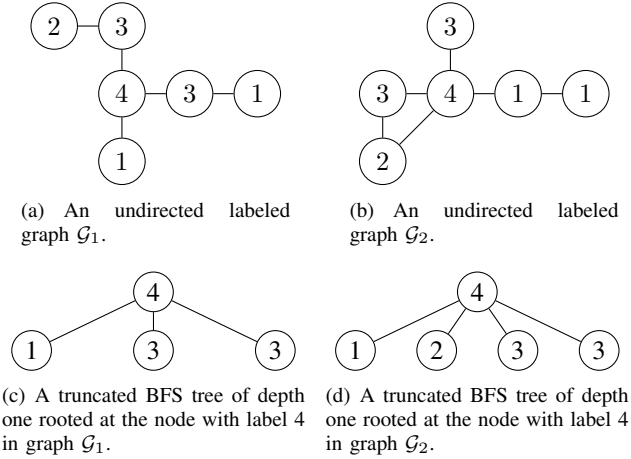

Guided by the small-world property, we construct a truncated BFS tree of depth $d$ rooted at each node to restrict the maximal length of the shortest path in a graph. Figure~\ref{fig:example}(a) and (b) show two undirected labeled graphs $\mathcal{G}_1$ and $\mathcal{G}_2$, respectively. Figure~\ref{fig:example}(c) and (d) show two truncated BFS trees of depth $d=1$ rooted at the nodes with label 4 in $\mathcal{G}_1$ and $\mathcal{G}_2$, respectively. All the shortest paths starting from the root of the BFS tree in Figure~\ref{fig:example}(c) can be represented by $``4", ``4,1", ``4,3", ``4,3"$ and those of the root of the BFS tree in Figure~\ref{fig:example}(d) can be represented by $``4", ``4,1", ``4,2", ``4,3", ``4,3"$. Note that we also consider the shortest path of length zero, which only contains one node that plays the roles of both source and sink node. We only consider the shortest paths in BFS trees instead of finding all the shortest paths in a graph as used in SP, considering the small-world property. For a set of graphs, we construct a truncated BFS tree of depth $d$ rooted at each node and generate all its shortest paths, whose representations are kept in a multiset\footnote{ A set that can contain the same element multiple times.} $\mathcal{M}$.

We define the shortest-path graph feature map as follows:
\begin{definition}[Shortest-path Graph Feature Map]
	\label{def:fmg}
	Let  $\mathcal{M}_1, \mathcal{M}_2, \ldots, \mathcal{M}_n$ be the multisets of shortest-path representations extracted from graphs $\mathcal{G}_1, \mathcal{G}_2, \ldots, \mathcal{G}_n$, respectively. Let the union of all the multisets be a set $\mathcal{U}=\mathcal{M}_1\cup\mathcal{M}_2\cup\cdots\cup\mathcal{M}_n=\{l(P_1), l(P_2),\cdots,l(P_{|\mathcal{U}|})\}$. Define a map $\psi: \{\mathcal{G}_1,\mathcal{G}_2,\ldots,\mathcal{G}_n\} \times \Sigma\rightarrow \mathbb{N}$ such that $\psi(\mathcal{G}_i, l(P_j))$ ($1\leq i\leq n, 1\leq j\leq |\mathcal{U}|$) is the frequency of the shortest path representation $l(P_j)$ in graph $\mathcal{G}_i$. Then, the shortest-path graph feature map of $\mathcal{G}_i$ is defined as follows:
	\begin{equation}
	\label{equ:graph}
	\phi(\mathcal{G}_i)=\left[ \psi(\mathcal{G}_i, l(P_1)),\psi(\mathcal{G}_i, l(P_2)),\ldots,\psi(\mathcal{G}_i, l(P_{|\mathcal{U}|}))\right]
	\end{equation}
\end{definition}

\subsection{Multi-scale Shortest-path Graph Feature Map}

In the last section, we decompose a graph into its substructures, i.e., shortest paths. However, shortest paths cannot reveal the structural information of nodes. Thus, the shortest-path graph feature map can only capture graph similarity at fine scales. To capture graph similarity at multiple different scales, we focus on the different scales of the graph structure around a node. 

To incorporate the multiple different scales of the graph structure into the shortest-path graph feature map, we first use the multiple different scales of the graph structure around a node to augment this node and then generate all the shortest paths starting from this node. Specifically, for each node in the shortest path $P=``v,v_1,v_2,\ldots,v'"$, we construct a truncated BFS tree of depth $k$ rooted at it, which captures the graph structure at the $k$-th scale around this node. And we have the shortest path $P$ that goes through a sequence of BFS trees $P=``\mathcal{T}_v,\mathcal{T}_{v_1},\mathcal{T}_{v_2},\ldots,\mathcal{T}_{v'}" $. Note that compared with the normal shortest path whose components are nodes, the shortest path in this work can contain BFS trees as its components.

Our next step is to represent such kinds of shortest paths. The difficulty is how to label each truncated BFS tree in the shortest path. Thus, we need to redefine the definition of the label function $l$ described in Section \ref{pre} as follows: $l: \mathcal{T}\rightarrow \Sigma$ is a function that assigns labels from a set of positive integers $\Sigma$ to BFS trees. Thus, we can just concatenate all the labels of BFS trees in the shortest path as its representation, i.e., $l(P)=``l(\mathcal{T}_v),l(\mathcal{T}_{v_1}),l(\mathcal{T}_{v_2}),\ldots, l(\mathcal{T}_{v'})" $.

\begin{figure}[htb!]
	\hspace*{\fill}
	\centering
	\subfigure[$\mathcal{T}_1^{(1)}$]{
		\begin{tikzpicture}
		[scale=0.9,level distance=8.66mm,
		every node/.style={draw, circle, minimum size=1.5em,inner sep=1},
		level 1/.style={sibling distance=12mm},
		level 2/.style={sibling distance=24mm}]
		\node {1}
		child {node {4}
			child {node {3}
				child[missing]
			}
			child {node {3}
				child[missing]
			}
		};
		\end{tikzpicture}
	}
	\hfill
	\centering
	\subfigure[$\mathcal{T}_2^{(1)}$]{
		\begin{tikzpicture}
		[scale=0.9,level distance=8.66mm,
		every node/.style={draw, circle, minimum size=1.5em,inner sep=1},
		level 1/.style={sibling distance=12mm},
		level 2/.style={sibling distance=24mm}]
		\node {1}
		child {node {3}
			child {node {4}
				child[missing]
			}
		};
		\end{tikzpicture}
	}
    \hspace*{\fill}
    
	\hspace*{\fill}
	\centering
	\subfigure[$\mathcal{T}_3^{(1)}$]{
		\begin{tikzpicture}
		[scale=0.9,level distance=8.66mm,
		every node/.style={draw, circle, minimum size=1.5em,inner sep=1},
		level 1/.style={sibling distance=12mm},
		level 2/.style={sibling distance=24mm}]
		\node {2}
		child {node {3}
			child {node {4}
				child[missing]
			}
		};
		\end{tikzpicture}
	}
	\hfill
	\centering
	\subfigure[$\mathcal{T}_4^{(1)}$]{
		\begin{tikzpicture}
		[scale=0.9,level distance=8.66mm,
		every node/.style={draw, circle, minimum size=1.5em,inner sep=1},
		level 1/.style={sibling distance=24mm},
		level 2/.style={sibling distance=12mm}]
		\node {3}
		child {node {2}
			child[missing]
		}
		child {node {4}
			child {node {1}
				child[missing]
			}
			child {node {3}
				child[missing]
			}
		};
		\end{tikzpicture}
	}
    \hspace*{\fill}
    
    \hspace*{\fill}
	\centering
	\subfigure[$\mathcal{T}_5^{(1)}$]{
		\begin{tikzpicture}
		[scale=0.9,level distance=8.66mm,
		every node/.style={draw, circle, minimum size=1.5em,inner sep=1},
		level 1/.style={sibling distance=24mm},
		level 2/.style={sibling distance=12mm}]
		\node {3}
		child {node {1}
			child[missing]
		}
		child {node {4}
			child {node {1}}
			child {node {3}}
		};
		\end{tikzpicture}
	}
	\hfill
	\centering
	\subfigure[$\mathcal{T}_6^{(1)}$]{
		\begin{tikzpicture}
		[scale=0.9,level distance=8.66mm,
		every node/.style={draw, circle, minimum size=1.5em,inner sep=1},
		level 1/.style={sibling distance=12mm},
		level 2/.style={sibling distance=12mm}]
		\node {4}
		child {node {1}
			child[missing]
		}
		child {node {3}
			child {node {1}}
		}
		child {node {3}
			child {node {2}}
		};
		\end{tikzpicture}
	}
	\hspace*{\fill}
	\caption{Truncated BFS trees of depth two rooted at each node in the undirected labeled graph $\mathcal{G}_1$.}
	\label{fig:truncatedtree1}
\end{figure}
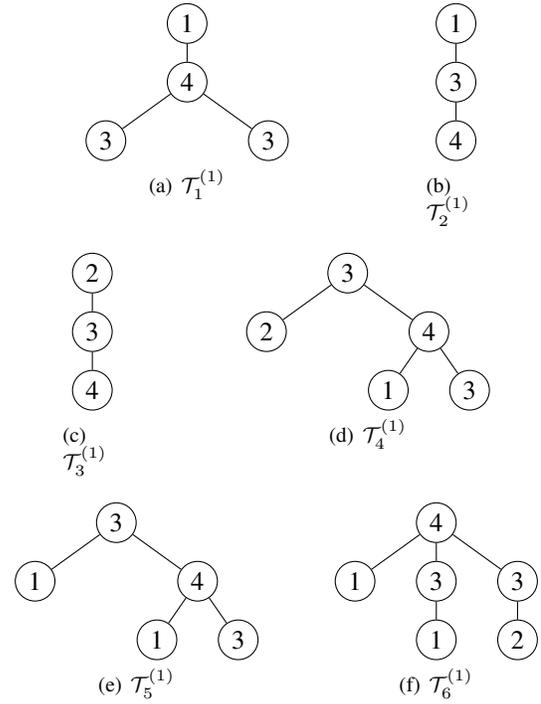

For each truncated BFS tree in the shortest path, we need to hash it to an integer label. The hash method should be injective, i.e., hashing the same two BFS trees to the same integer label and different BFS trees to different integer labels. In this paper, we propose a method for this purpose. We use Figure~\ref{fig:truncatedtree1}(f) as an example, which shows a truncated BFS tree of depth two rooted at the node with label 4 in graph $\mathcal{G}_1$. We recursively consider each subtree of depth one in this truncated BFS tree. Firstly, we consider the subtree of depth one rooted at the node with label 4 (also shown in Figure~\ref{fig:example}(c)). We sort the child nodes in ascending order with regard to their label values and use the concatenation of all the edges in the subtree as its representation. Thus, this subtree can be represented by a string $``4,1,4,3,4,3"$. Likewise, the middle subtree of depth one rooted at the node with label 3 in Figure~\ref{fig:truncatedtree1}(f) can be represented by a string $``3,1"$ and the right subtree of depth one rooted at the node with label 3 in Figure~\ref{fig:truncatedtree1}(f) can be represented by a string $``3,2"$. Note that the roots of the latter two subtrees of depth one have the same label, which leads to arbitrariness in node order. To eliminate arbitrariness in node order, we sort the subtrees of depth one rooted at the nodes with the same label lexicographically according to their string representations. Finally, the subtree $\mathcal{T}_6^{(1)}$ as shown in Figure~\ref{fig:truncatedtree1}(f) can be represented by a string $``4,1,4,3,4,3-3,1-3,2"$, which is the concatenation of the representations of all the subtrees of depth one in it. We use the dash symbol $``-"$ to differentiate between different subtrees. This dash symbol is necessary to differentiate $\mathcal{T}_6^{(1)}$ between the case in which the node with label 2 is attached to the root of the middle subtree of depth one rooted at the node with label 3. Our method is injective and makes the representation of each truncated BFS tree unique. Now, the label function $l: \mathcal{T}\rightarrow \Sigma$ can assign the same positive integer label to the same trees, which have the same representation. 

\begin{figure}[htb!]
	\hspace*{\fill}
	\centering
	\subfigure[$\mathcal{T}_1^{(2)}$]{
		\begin{tikzpicture}
		[scale=0.9,level distance=8.66mm,
		every node/.style={draw, circle, minimum size=1.5em,inner sep=1},
		level 1/.style={sibling distance=12mm},
		level 2/.style={sibling distance=12mm}]
		\node {1}
		child {node {1}
			child {node {4}
				child[missing]
			}
		};
		\end{tikzpicture}
	}
	\hfill
	\centering
	\subfigure[$\mathcal{T}_2^{(2)}$]{
		\begin{tikzpicture}
		[scale=0.9,level distance=8.66mm,
		every node/.style={draw, circle, minimum size=1.5em,inner sep=1},
		level 1/.style={sibling distance=24mm},
		level 2/.style={sibling distance=12mm}]
		\node {1}
		child {node {1}
			child[missing]
		}
		child{node {4}
			child {node {2}
				child[missing]
			}
		    child {node {3}
		    	child[missing]
		    }
	        child {node {3}
	        	child[missing]
	        }
		};
		\end{tikzpicture}
	}
    \hspace*{\fill}
    
    \hspace*{\fill}
	\centering
	\subfigure[$\mathcal{T}_3^{(2)}$]{
		\begin{tikzpicture}
		[scale=0.9,level distance=8.66mm,
		every node/.style={draw, circle, minimum size=1.5em,inner sep=1},
		level 1/.style={sibling distance=24mm},
		level 2/.style={sibling distance=12mm}]
		\node {2}
		child {node {3}
				child[missing]
			}
		child {node {4}
			child {node {1}
				child[missing]
			}
		    child {node {3}
		    	child[missing]
		    }
		};
		\end{tikzpicture}
	}
	\hfill
	\centering
	\subfigure[$\mathcal{T}_4^{(2)}$]{
		\begin{tikzpicture}
		[scale=0.9,level distance=8.66mm,
		every node/.style={draw, circle, minimum size=1.5em,inner sep=1},
		level 1/.style={sibling distance=12mm},
		level 2/.style={sibling distance=12mm}]
		\node {3}
		child {node {4}
			child {node {1}
				child[missing]
			}
		    child {node {2}
		    	child[missing]
		    }
	        child {node {3}
	        	child[missing]
	        }
		};
		\end{tikzpicture}
	}
    \hspace*{\fill}
    
    \hspace*{\fill}
	\centering
	\subfigure[$\mathcal{T}_5^{(2)}$]{
		\begin{tikzpicture}
		[scale=0.9,level distance=8.66mm,
		every node/.style={draw, circle, minimum size=1.5em,inner sep=1},
		level 1/.style={sibling distance=24mm},
		level 2/.style={sibling distance=12mm}]
		\node {3}
		child {node {2}
			child[missing]
		}
		child {node {4}
			child {node {1}}
			child {node {3}}
		};
		\end{tikzpicture}
	}
	\hfill
	\centering
	\subfigure[$\mathcal{T}_6^{(2)}$]{
		\begin{tikzpicture}
		[scale=0.9,level distance=8.66mm,
		every node/.style={draw, circle, minimum size=1.5em,inner sep=1},
		level 1/.style={sibling distance=8mm},
		level 2/.style={sibling distance=8mm}]
		\node {4}
		child {node {1}
			child {node {1}
				child[missing]
			}
		}
	    child {node {2}
	    	child[missing]
	    }
	    child {node {3}
	    	child[missing]
	    }
        child {node {3}
        	child[missing]
        };
		\end{tikzpicture}
	}
	\hspace*{\fill}
	\caption{Truncated BFS trees of depth two rooted at each node in the undirected labeled graph $\mathcal{G}_2$.}
	\label{fig:truncatedtree2}
\end{figure}
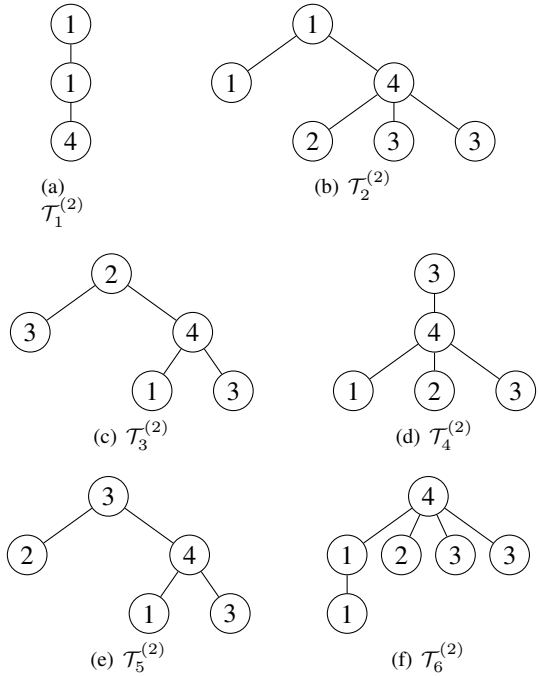

All unique BFS trees of depth $k$ rooted at each node in a dataset of graphs are kept in a set. For example, For BFS trees shown in Figure~\ref{fig:truncatedtree1} and Figure~\ref{fig:truncatedtree2}, the set will contain $\mathcal{T}_1^{(2)}$, $\mathcal{T}_2^{(2)}$, $\mathcal{T}_2^{(1)}$, $\mathcal{T}_1^{(1)}$, $\mathcal{T}_3^{(2)}$, $\mathcal{T}_3^{(1)}$, $\mathcal{T}_5^{(1)}$, $\mathcal{T}_4^{(1)}/\mathcal{T}_5^{(2)}$, $\mathcal{T}_4^{(2)}$, $\mathcal{T}_6^{(2)}$, and $\mathcal{T}_6^{(1)}$. All unique BFS trees in the set are sorted lexicographically. The index of each truncated BFS tree in the set can be used as its label. For instance,  $l: \mathcal{T}_1^{(2)}\rightarrow 1$, $l: \mathcal{T}_2^{(2)}\rightarrow 2$, $l: \mathcal{T}_2^{(1)}\rightarrow 3$, $l: \mathcal{T}_1^{(1)}\rightarrow 4$, $l: \mathcal{T}_3^{(2)}\rightarrow 5$, $l: \mathcal{T}_3^{(1)}\rightarrow 6$, $l: \mathcal{T}_5^{(1)}\rightarrow 7$, $l: \mathcal{T}_4^{(1)}/\mathcal{T}_5^{(2)}\rightarrow 8$, $l: \mathcal{T}_4^{(2)} \rightarrow 9$, $l: \mathcal{T}_6^{(2)} \rightarrow 10$, and $l: \mathcal{T}_6^{(1)} \rightarrow 11$. After considering the graph structure at the second scale around node 4 in $\mathcal{G}_1$, the shortest paths starting from node 4 are $``\mathcal{T}_6^{(1)}"$, $``\mathcal{T}_6^{(1)}, \mathcal{T}_1^{(1)}"$, $``\mathcal{T}_6^{(1)}, \mathcal{T}_4^{(1)}"$, and $``\mathcal{T}_6^{(1)}, \mathcal{T}_5^{(1)}"$. Their corresponding representations are $``11"$, $``11, 4"$, $``11, 8"$, and $``11, 7"$, respectively. 

Note that if the truncated BFS trees of the same depth rooted at two nodes are identical, the two nodes have the same scale of the graph structure around them. For example, the two nodes with label 3 on the upper side of the node with label 4 in Figure~\ref{fig:example}(a) and on the left side of the node with label 4 in Figure~\ref{fig:example}(b) have the same BFS trees of depth two as shown in Figure~\ref{fig:truncatedtree1}(d) and Figure~\ref{fig:truncatedtree2}(e), respectively. Thus, they have the same second scale of the graph structure. Figure~\ref{fig:truncatedtree1}(a) and (b) show another two truncated BFS trees rooted at the two nodes with the same label 1 in Figure~\ref{fig:example}(a). We can see that they have a different second scale of the graph structure. Thus, by integrating the multiple different scales of the graph structure into the shortest paths, we can distinguish graphs at multiple different scales. If we build truncated BFS trees of depth zero rooted at each node for augmenting nodes, the two shortest paths that have the same representation $``1,4,3"$ in Figure~\ref{fig:example}(a) (The starting node is the node with label 1 on the bottom side of the node with label 4 and the ending node is the node with label 3 on the upper side of the node with label 4.) and Figure~\ref{fig:example}(b) (The starting node is the node with label 1 on the right side of the node with label 4 and the ending node is the node with label 3 on the left side of the node with label 4.) cannot be distinguished. However, if we augment nodes using truncated BFS trees of depth two (as shown in Figure~\ref{fig:truncatedtree1}(a), (f), and (d) and Figure~\ref{fig:truncatedtree2}(b), (f), and (e), respectively), the aforementioned two shortest paths have different representations $``4,11,8"$ and $``2,10,8"$, which can be distinguished easily.

In this paper, we concatenate all the shortest-path graph feature maps at multiple different scales as follows:
\begin{equation}
\label{equ:multiscaleG}
\phi(\mathcal{G})=\phi(\mathcal{G}^{(0)})\parallel\phi(\mathcal{G}^{(1)})\parallel\cdots\parallel\phi(\mathcal{G}^{(k)})
\end{equation}
where $\phi(\mathcal{G}^{(k)})$ corresponds to the shortest-path graph feature map built on the shortest paths with truncated BFS trees of depth $k$ rooted at each node as its components.

\subsection{Computing Graph Kernels}

We define the shortest-path node feature map as follows:
\begin{definition}[Shortest-path Node Feature Map]
	\label{def:fmv}
	For nodes in graph $\mathcal{G}$, define a map $\psi: \{v_1,v_2,\ldots,v_{|\mathcal{V}|}\} \times \Sigma \rightarrow \mathbb{N}$ where $v_i\in \mathcal{V} (1\leq i\leq|\mathcal{V}|)$ such that $\psi(v_i, l(P_j))$ ($1\leq i\leq |\mathcal{V}|, 1\leq j\leq |\mathcal{U}|$) is the frequency of the shortest path $l(P_j)$ that starts from $v_i$ in graph $\mathcal{G}$. Then, the shortest-path node feature map of $v_i$ is defined as follows:
	\begin{equation}
	\label{equ:vertex}
	\phi(v_i)=\left[ \psi(v_i, l(P_1)),\psi(v_i, l(P_2)),\ldots,\psi(v_i, l(P_{|\mathcal{U}|}))\right]
	\end{equation}
\end{definition}

Similar to Equation~(\ref{equ:multiscaleG}), we generate the multi-scale shortest-path node feature map as follows:
\begin{equation}
\label{equ:multiscaleV}
\phi(v_i)=\phi(v_i^{(0)})\parallel\phi(v_i^{(1)})\parallel\cdots\parallel\phi(v_i^{(k)})
\end{equation}
where $\phi(v_i^{(k)})$ is the shortest-path node feature map built on $\mathcal{G}^{(k)}$.

We can see that the multi-scale shortest-path graph feature map equals the sum of all the multi-scale shortest-path node feature maps in that graph, i.e.,
\begin{equation}
\label{equ:add}
\phi(\mathcal{G})=\sum_{i=1}^{|\mathcal{V}|}\phi(v_i)
\end{equation}

In this paper, we use the multi-scale shortest-path node feature maps rather than their sum aggregation, i.e., the multi-scale shortest-path graph feature map, because the sum aggregation loses the distributions of shortest paths around each node. We first use Equation~(\ref{equ:w1}) to compute the distance between the multi-scale shortest-path node feature maps of two graphs $\mathcal{G}_1$ and $\mathcal{G}_2$ and then use the Laplacian RBF kernel to construct our graph kernel MWSP:
\begin{equation}
\label{equ:mwsp}
\kappa(\mathcal{G}_1, \mathcal{G}_2)=\exp\left(-\lambda W_1(\mathbf{X},\mathbf{Y})\right)
\end{equation}
where $\lambda$ is a hyperparameter and $\mathbf{X}$ and $\mathbf{Y}$ denote the matrix of the multi-scale shortest-path node feature maps of graphs $\mathcal{G}_1$ and $\mathcal{G}_2$, respectively.

Note that the Wasserstein distance, despite being a metric, in its general form is not isometric (i.e., no metric-preserving mapping to the $l_2$ norm) as argued in~\cite{togninalli2020wasserstein}, owing to that the metric space induced by it strongly depends on the chosen ground distance~\cite{figalli2011optimal}. Thus, we cannot derive a positive semi-definite kernel from the Wasserstein distance in its general form, because the conventional methods~\cite{haasdonk2004learning} cannot be applied to it. Thus, MWSP is not a positive semi-definite kernel. However, we can still employ Kre{\u{\i}}n SVM (KSVM)~\cite{loosli2015learning} for learning with indefinite kernels. Generally speaking, kernels that are not positive semi-definite induce reproducing kernel Kre{\u{\i}}n spaces (RKKS), which is a generalization of reproducing kernel Hilbert spaces (RKHS). KSVM can solve learning problems in RKKS.

\subsection{Algorithm}\label{complex}

In this section, we provide the pseudo-code of our MWSP graph kernel in Algorithm~1. Line 2 in Algorithm~1 uses Algorithm~2 to generate all the shortest paths in a set of graphs $\mathcal{G}$. For each node $v$ in graph $\mathcal{G}_i$ (lines 4--9 in Algorithm~2), we build a truncated BFS tree of depth $d$ rooted at it. The time complexity of the BFS traversal of a graph is $\mathcal{O}\left(|\mathcal{V}|+|\mathcal{E}|\right)$, where $|\mathcal{V}|$ is the number of nodes and $|\mathcal{E}|$ is the number of edges in a graph. For convenience, we assume that $|\mathcal{E}|>|\mathcal{V}|$ and all the $n$ graphs have the same number of nodes and edges. Thus, the time complexity of Algorithm~2 for all the $n$ graphs is $\mathcal{O}\left( n|\mathcal{V}|\left(  |\mathcal{E}|+|\mathcal{V}|\right) \right) $. Line 3 in Algorithm~1 uses Algorithm~3 to augment/relabel each node (i.e., hashing each truncated BFS tree of depth $k$ to an integer label, capturing the multiple different scales of the graph structure). Like Algorithm~2, for each node in each graph, Algorithm~3 also constructs a BFS tree of depth $k$, whose time complexity is $\mathcal{O}\left( kn|\mathcal{V}|\left(  |\mathcal{E}|+|\mathcal{V}|\right) \right) $. For a total of $kn|\mathcal{V}|$ truncated BFS trees, we use our hash method to project them to unique integer labels, whose time complexity is bounded by $\mathcal{O}\left(n|\mathcal{E}|\log\left(|\mathcal{E}|\right)\right)$. Thus, the time complexity of Algorithm~3 is $\mathcal{O}\left( kn|\mathcal{V}|\left(  |\mathcal{E}|+|\mathcal{V}|\right)+ n|\mathcal{E}|\log\left(|\mathcal{E}|\right)\right) $.

Lines 6--10 in Algorithm~1 generate all the shortest paths corresponding to the same value of $k$ (i.e., the same scale of the graph structure). The number of shortest paths generated by Algorithm~2 is at most $n|\mathcal{V}|^2 $. The time complexity of lines 6--10 in Algorithm~1 is $\mathcal{O}\left(nd|\mathcal{V}|^2\right) $ since the depth of the truncated BFS tree or the maximal length of all the shortest paths is $d$. Line 11 in Algorithm~1 sorts the elements in $\mathcal{U}^{(i)}$ lexicographically. Since $\mathcal{U}^{(i)}$ at most contains $|\mathcal{V}|^2$ shortest paths in terms of their representations. The time complexity is bounded by $\mathcal{O}\left(d|\mathcal{V}|^2\right)$ via radix sort. Lines 12--15 in Algorithm~1 count the number of occurrences of each unique shortest path in $\mathcal{U}^{(i)}$ around each node. For each node, the counting time complexity is bounded by $\mathcal{O}\left(qd|\mathcal{V}|\right)$, where $q$ is the maximal length of all $\mathcal{U}^{(i)} (0\leq i\leq k)$. Thus, the time complexity of lines 12--15 in Algorithm~1 is $\mathcal{O}\left(nqd|\mathcal{V}|^2\right)$. The time complexity of lines 4--15 in Algorithm~1 is $\mathcal{O}\left(knqd|\mathcal{V}|^2\right)$.

Lines 18--19 in Algorithm~1 first compute the Wasserstein distance between the multi-scale shortest-path node feature maps of graphs $\mathcal{G}_i$ and $\mathcal{G}_j$ and then compute the Laplacian RBF kernel. The time complexity of computing the Wasserstein distance is $\mathcal{O}\left(|\mathcal{V}|^3\log(|\mathcal{V}|)\right)$~\cite{togninalli2019wasserstein}. For $n$ graphs, the time complexity to compute the kernel matrix is $\mathcal{O}\left(n^2|\mathcal{V}|^3\log(|\mathcal{V}|)\right)$. The total time complexity of our MWSP graph kernel is bounded by $\mathcal{O}\left(kn|\mathcal{V}|\left(|\mathcal{E}|+qd|\mathcal{V}|\right)+n|\mathcal{E}|\log\left(|\mathcal{E}|\right)+n^2|\mathcal{V}|^3\log(|\mathcal{V}|)\right)$.

\begin{algorithm2e}
	\SetKwFunction{pg}{Shortest-path\_Generation}
	\SetKwFunction{gr}{Graph\_Augmentation}
	\KwIn{A set of graphs $\mathcal{G}=\{\mathcal{G}_1, \mathcal{G}_2, \ldots, \mathcal{G}_n\}$ and the initial label function $l: \mathcal{V}\rightarrow \Sigma$, $d$, $k$, $\lambda$}
	\KwOut{The computed kernel matrix $\mathbf{K}\in\mathbb{R}^{n\times n}$}
	$\mathbf{K}\leftarrow$ \upshape zeros(($n$,$n$))\;
	all\_graph\_paths$\leftarrow$\pg$\left(\mathcal{G},d\right)$\;
	all\_labels$\leftarrow$\gr$\left(\mathcal{G},l,k\right)$\;
	\For{$i \leftarrow 0$ \KwTo $k$ }{
		$\mathcal{U}^{(i)}\leftarrow $\upshape set( ), $l^{(i)}\leftarrow $all\_labels[$i$]\;
		\For{$j \leftarrow 1$ \KwTo $n$ }{
			graph\_path$\leftarrow$ all\_graph\_paths[$j$]\;
			\ForEach{\upshape node\_path$\in$ \upshape graph\_path}{
				\ForEach{\upshape path $P\in$ node\_path }{
					$\mathcal{U}^{(i)}$.add($l^{(i)}(P)$)\;
				}
			}
		}
		$\mathcal{U}^{(i)}\leftarrow $\upshape sort($\mathcal{U}^{(i)}$)\tcc*[r]{sort lexicographically}
		\For{$j \leftarrow 1$ \KwTo $n$ }{
			\ForEach{\upshape node $v \in$ \upshape graph $\mathcal{G}_j$}{
			    $\phi(v)\leftarrow [ \psi\left(v,l^{(i)}(P_1)\right), \psi\left(v,l^{(i)}(P_2)\right),\ldots$
			    $\ldots,\psi\left(v,l^{(i)}(P_{|\mathcal{U}^{(i)}|})\right)] $\;
		    }
			$\mathbf{\Phi}_j^{(i)}\leftarrow\left[ \phi(v_1);\phi(v_2);\ldots;\phi(v_{|\mathcal{V}_j|})\right] $\tcc*[r]{row-wise concatenation}
			
		}
	}
	\For{$j \leftarrow 1$ \KwTo $n$ }{
		$\mathbf{\Phi}_j\leftarrow\left[\mathbf{\Phi}_j^{(0)}\parallel\mathbf{\Phi}_j^{(1)}\parallel\cdots\parallel\mathbf{\Phi}_j^{(k)}\right] $\tcc*[r]{column-wise concatenation}	
	}
	\ForEach{\upshape pair of graphs $\mathcal{G}_i$ \upshape and $\mathcal{G}_j$}{
		$\mathbf{K}(i, j)\leftarrow\exp\left(-\lambda W_1(\mathbf{\Phi}_i,\mathbf{\Phi}_j)\right)$
	}
	\Return{$\mathbf{K}$} \;
	\caption{MWSP}
	\label{alg:mwsp}
\end{algorithm2e}

\begin{algorithm2e}
	\KwIn{A set of graphs $\{\mathcal{G}_1, \mathcal{G}_2, \ldots, \mathcal{G}_n\}$ and the depth $d$ of the truncated BFS tree rooted at each node}
	\KwOut{All the shortest paths in these graphs}
	all\_graph\_paths$\leftarrow $[]\;
	\For{$i \leftarrow 1$ \KwTo $n$ }{
		graph\_paths$\leftarrow $[]\;
		\ForEach{\upshape node $v\in\mathcal{G}_i$ }{
			node\_paths$\leftarrow $[]\;
			Build a truncated BFS tree $\mathcal{T}_v$ of depth $d$ rooted at node $v$\;
			\ForEach{\upshape node $v'\in\mathcal{T}_v$ }{
				node\_paths.append($P$)\tcc*[r]{$P=``v,v_1,v_2,\ldots,v'"$}
			}
			graph\_paths.append(node\_paths)\;
		}
		all\_graph\_paths.append(graph\_paths)\;
	}
	
	\Return{\upshape all\_graph\_paths} \;
	\caption{Shortest-path\_Generation}
	\label{alg:pg}
\end{algorithm2e}

\begin{algorithm2e}
	\KwIn{A set of graphs $\mathcal{G}=\{\mathcal{G}_1, \mathcal{G}_2, \ldots, \mathcal{G}_n\}$ and the initial label function $l: \mathcal{T}\rightarrow \Sigma$, the depth $k$ of the truncated BFS tree rooted at each node for capturing the multiple different scales of the graph structure}
	\KwOut{The new label of each node}
	all\_labels$\leftarrow $[]\;
	\For{$i \leftarrow 0$ \KwTo $k$ }{
		$l^{(i)}\leftarrow $[]\;
		\For{$j \leftarrow 1$ \KwTo $n$ }{
			label$\leftarrow $[]\;
			\ForEach{\upshape node $v\in\mathcal{G}_j$ }{
				Build a truncated BFS tree $\mathcal{T}_{v}$ of depth $i$ rooted at node $v$\;
				label.append($l(\mathcal{T}_{v})$)\;
			}
			$l^{(i)}$.append(label)
		}
		all\_labels.append($l^{(i)}$)\;
	}
	\Return{\upshape all\_labels} \;
	\caption{Graph\_Augmentation}
	\label{alg:gr}
\end{algorithm2e}

\section{Experimental Evaluation}\label{experiments}
\subsection{Experimental Setup}
We run all the experiments on a server with a dual-core Intel(R) Xeon(R) Gold 6226R CPU @ 2.90GHz, 256 GB memory, and Ubuntu 18.04.6 LTS operating system. MWSP is written in Python. We make our code publicly available at Github\footnote{\url{https://github.com/yeweiysh/MWSP}}. We compare MWSP with 9 state-of-the-art graph kernels, i.e., MLG~\cite{kondor2016multiscale}, RetGK~\cite{zhang2018retgk}, SP~\cite{borgwardt2005shortest}, WL~\cite{shervashidze2011weisfeiler}, FWL~\cite{schulz2021graph}, WWL~\cite{togninalli2020wasserstein}, GWL~\cite{bause2022gradual}, GAWL~\cite{nikolentzos2023graph}, and GraphQNTK~\cite{tanggraphqntk}. 

We set the parameters for our MWSP graph kernel as follows: The depth $d$ of the truncated BFS tree rooted at each node for constructing and limiting the maximal length of the shortest paths is chosen from \{0, 1, 2, \ldots, 6\} and the depth $k$ of the truncated BFS tree for capturing the graph structure at multiple different scales around each node is chosen from \{0, 1, 2, \ldots, 6\} as well. Both $d$ and $k$ are selected by cross-validation on the training data. For a specific value of $k$ or $d$, if MWSP runs out of memory on some dataset, we will not try larger values and terminate the search. The parameters of the comparison methods are set according to their original papers. We use the implementations of MLG, SP, and WL from the GraKeL~\cite{siglidis2018grakel} library. The implementations of other methods are obtained from their official websites. 

We use 10-fold cross-validation with a binary $C$-SVM \cite{chang2011libsvm} (or a KSVM~\cite{loosli2015learning} for MWSP, FWL, and WWL) to test the classification accuracy of each method. The parameter $C$ for each fold is independently tuned from $\left\lbrace10^{-3}, 10^{-2}, \ldots,10^{3}\right\rbrace $ using the training data from that fold. We also tune the parameter $\lambda$ from $\left\lbrace10^{-4}, 10^{-3}, \ldots,10^{1}\right\rbrace $ for MWSP, FWL, and WWL. We repeat the experiments 10 times and report the average classification accuracy and standard deviation.

In order to test the performance of our graph kernel MWSP, we adopt 14 benchmark datasets whose statistics are given in Table \ref{tab:statistics_dataset}. All the datasets are downloaded from Kersting et al.~\cite{KKMMN2016}. Chemical compound datasets include MUTAG~\cite{debnath1991structure}, DHFR\_MD~\cite{sutherland2003spline}, NCI1~\cite{wale2008comparison}, and NCI109~\cite{wale2008comparison}. These chemical compounds are represented by graphs, of which edges represent the chemical bond types (single, double, triple, or aromatic), nodes represent atoms, and node labels indicate atom types. Molecular compound datasets contain PTC (male mice (MM), male rats (MR), female mice (FM) and female rats (FR))~\cite{kriege2012subgraph}, ENZYMES~\cite{borgwardt2005protein}, PROTEINS~\cite{borgwardt2005protein}, and DD~\cite{dobson2003distinguishing}. IMDB-B~\cite{yanardag2015deep} is a movie collaboration dataset, which contains movies of different actors/actresses and genre information. Graphs represent the collaborations between different actors/actresses. For each actor/actress, a corresponding collaboration graph (ego network) is derived and labeled with its genre. Nodes represent actors/actresses and edges denote that two actors/actresses appear in the same movie. The collaboration graphs are generated on the Action and Romance genres. REDDIT-B~\cite{yanardag2015deep} contains graphs representing online discussion threads. Nodes represent users and edges represent the responses between users. This dataset has two classes, i.e., question/answer-based community and discussion-based community. Cuneiform dataset~\cite{kriege2018recognizing} contains graphs representing 30 different Hittite cuneiform signs, each of which consists of tetrahedron-shaped markings called wedges. The visual appearance of wedges can be classified according to the stylus orientation into vertical, horizontal, and diagonal. Each type of wedge is represented by a graph consisting of four nodes, i.e., right node, left node, tail node, and depth node. 

\begin{table*}[!htb]
	\centering
	\caption{Statistics of the benchmark datasets used in the experiments. Mean SP length stands for the average length of all the shortest paths in a graph while Max SP length stands for the maximal length of all the shortest paths in a graph. N/A means a graph has no node labels.}
	\label{tab:statistics_dataset}
	\begin{tabular}{l|l|l|l|l|l|l|l}
		\toprule
		Dataset         &Size            & Class \#  & Average node \#    & Average edge \#  &Node label \# &Mean SP length&Max SP length\\ \hline
		MUTAG           &188         &2              &17.93             &19.79            &7 &3.87&15\\
		DHFR\_MD      &393       &2               &23.87            &283.01         &7               &1.00&1\\
		NCI1                  &4110       &2              &29.87            &32.30           &37         &7.27&45\\
		NCI109            &4127        &2             &29.68             &32.13            &38         &7.19&61\\
		PTC\_MM        &336         &2             &13.97             &14.32            &20             &5.34&30\\
		PTC\_MR         &344          &2            &14.29             &14.69            &18               &5.70&30\\
		PTC\_FM          &349         &2             &14.11              &14.48            &18               &5.23&30\\
		PTC\_FR          &351          &2             &14.56             &15.00            &19               &5.54&30\\
		ENZYMES        &600          &6           &32.63              &62.14           &3            &5.72&37\\
		PROTEINS         &1113        &2           &39.06             &72.82           &3         &10.55&64\\
		DD                     &1178         &2           &284.32           &715.66          &82&16.84&83\\
		IMDB-B              &1000        &2          &19.77              &96.53           &N/A&1.59&2\\     
		REDDIT-B           &2000        &2          &429.63         &497.75           &N/A&3.38&19\\  
		Cuneiform        &267          &30         &21.27             &44.80            &3&2.30&3\\
		\bottomrule
	\end{tabular}
\end{table*}

\begin{table*}[!htb]
	\centering
	\caption{Comparison of classification accuracy (mean $\pm$ standard deviation) of MWSP to other graph kernels on the benchmark datasets. N/A means the results are not reproducible because of running out of time (48h for MLG) or because the current code of FWL does not support that specific dataset. $^\dag$ means the results are adopted from the original papers.}
	\label{tab:classification}
	\begin{tabular}{l|l|l|l|l|l|l|l|l|l|l}
		\toprule
		Dataset             &MWSP                                  &MLG                                 &RetGK                                            &SP                       &WL                                     &FWL                     &WWL                               &GWL&GAWL&GraphQNTK\\ \hline
		MUTAG             &89.9$\pm$5.0                    &84.2$\pm$2.6$^\dag$  &\textbf{90.3$\pm$1.1}$^\dag$ &84.4$\pm$1.7    &82.1$\pm$0.4$^\dag$   &85.7$\pm$7.5    &87.3$\pm$1.5$^\dag$ &86.0$\pm$1.2&87.3$\pm$6.3$^\dag$ &88.4$\pm$6.5$^\dag$     \\
		DHFR\_MD       &\textbf{73.5$\pm$5.2}      &67.1$\pm$0.6                   &64.4$\pm$1.0                              &68.0$\pm$0.4   &64.0$\pm$0.5                 &64.4$\pm$5.6    &67.9$\pm$1.1                 &64.9$\pm$0.9&67.9$\pm$2.5&67.9$\pm$8.4   \\
		NCI1                   &\textbf{86.3$\pm$1.1}     &80.8$\pm$1.3$^\dag$     &84.5$\pm$0.2$^\dag$                &73.1$\pm$0.3   &82.2$\pm$0.2$^\dag$   &85.4$\pm$1.6    &85.8$\pm$0.3$^\dag$ &85.3$\pm$0.4$^\dag$ &85.9$\pm$1.2$^\dag$ &77.2$\pm$2.7$^\dag$   \\
		NCI109              &\textbf{86.4$\pm$1.4}      &81.3$\pm$0.8$^\dag$    &80.1$\pm$0.5                               &72.8$\pm$0.3   &82.5$\pm$0.2$^\dag$   &85.8$\pm$0.9    &86.3$\pm$1.5                &81.7$\pm$0.2&85.4$\pm$1.3&72.7$\pm$2.5    \\
		PTC\_MM         &\textbf{71.1$\pm$6.5}        &61.2$\pm$1.1                    &67.9$\pm$1.4$^\dag$                 &62.2$\pm$2.2   &67.2$\pm$1.6                  &67.0$\pm$5.1     &69.1$\pm$5.0                &65.1$\pm$1.8&66.3$\pm$5.3&66.4$\pm$12.7  \\
		PTC\_MR          &\textbf{67.1$\pm$4.0}      &62.1$\pm$2.2                    &62.5$\pm$1.6$^\dag$                  &59.9$\pm$2.0  &61.3$\pm$0.9                 &59.3$\pm$7.3     &66.3$\pm$1.2$^\dag$ &59.9$\pm$1.6&59.0$\pm$4.4&59.4$\pm$9.9    \\
		PTC\_FM           &\textbf{67.7$\pm$4.1}     &61.1$\pm$1.8                    &63.9$\pm$1.3$^\dag$                &61.4$\pm$1.7     &64.4$\pm$2.1                 &61.0$\pm$6.9      &65.3$\pm$6.2               &62.6$\pm$1.9$^\dag$ &63.9$\pm$4.4&63.8$\pm$5.3   \\
		PTC\_FR            &\textbf{70.1$\pm$5.3}      &63.7$\pm$1.9                    &67.8$\pm$1.1$^\dag$                &66.9$\pm$1.5    &66.2$\pm$1.0                   &67.2$\pm$4.6     &67.3$\pm$4.2               &66.0$\pm$1.1&64.7$\pm$3.6&68.9$\pm$9.0  \\
		ENZYMES         &\textbf{66.7$\pm$4.7}      &57.9$\pm$5.4$^\dag$     &60.4$\pm$0.8$^\dag$              &41.1$\pm$0.8     &52.2$\pm$1.3$^\dag$     &51.8$\pm$5.5     &59.1$\pm$0.8$^\dag$ &54.5$\pm$1.6&58.5$\pm$4.8&35.8$\pm$5.3   \\
		PROTEINS        &\textbf{76.6$\pm$3.4}       &76.1$\pm$2.0$^\dag$     &75.8$\pm$0.6$^\dag$                &75.8$\pm$0.6   &75.5$\pm$0.3                  &74.6$\pm$3.8     &74.3$\pm$0.6$^\dag$  &73.7$\pm$0.5&74.7$\pm$3.0$^\dag$ &71.1$\pm$3.2$^\dag$   \\
		DD                      &\textbf{80.3$\pm$3.3}                   &74.6$\pm$0.8                                     &79.9$\pm$0.7   &79.3$\pm$0.3   &79.8$\pm$0.4$^\dag$  &78.4$\pm$2.4     &79.7$\pm$0.5$^\dag$  &79.0$\pm$0.8$^\dag$&78.7$\pm$2.8$^\dag$ &79.6$\pm$4.7 \\
		IMDB-B            &\textbf{75.2$\pm$2.7}       &68.1$\pm$1.1                    &72.3$\pm$0.6$^\dag$                  &72.2$\pm$0.8   &73.8$\pm$3.9                  &72.0$\pm$4.6    &73.3$\pm$4.1                   &73.7$\pm$1.3$^\dag$ &74.5$\pm$4.1$^\dag$ &73.3$\pm$3.6$^\dag$  \\
		REDDIT-B        &90.8$\pm$1.2       &N/A                    &\textbf{91.5$\pm$0.2}                &84.6$\pm$0.2   &81.0$\pm$3.1                 &79.0$\pm$2.2    &86.4$\pm$0.9                   &86.5$\pm$0.4$^\dag$ &88.0$\pm$1.4$^\dag$&84.8$\pm$1.7  \\
		Cuneiform         &\textbf{79.8$\pm$0.7}        &73.1$\pm$1.1                                     &73.3$\pm$1.1                                 &74.4$\pm$0.9   &73.4$\pm$0.9                   &N/A                      &74.4$\pm$1.7                  &74.0$\pm$1.1&73.2$\pm$1.3&78.1$\pm$3.5  \\
		\bottomrule
	\end{tabular}
\end{table*}

\subsection{Results}

\subsubsection{Classification Accuracy}

Table~\ref{tab:classification} demonstrates the comparison of the classification accuracy of MWSP to other graph kernels on the benchmark datasets. Our method MWSP needs node labels. If datasets (e.g., social networks) do not have node labels, we follow the method described in~\cite{xu2018powerful}, i.e., assigning node degrees as node labels for the IMDB-B dataset and setting uniform node labels for the REDDIT-B dataset. 
We can see that MWSP achieves the best classification accuracy on 12 datasets. 
On the ENZYMES dataset, MWSP has a gain of 10.4\% over the best comparison method RetGK and a gain of 86.3\% over the worst comparison method GraphQNTK. MWSP is outperformed by RetGK on two datasets, i.e., MUTAG and REDDIT-B. But RetGK is not an R-convolution graph kernel. It uses the return probability of random walk starting from and ending in the same node with multiple hops as node features and adopts the Maximum Mean Discrepancy (MMD)~\cite{gretton2012kernel} to compute graph similarity. Compared with all the R-convolution graph kernels, MWSP performs the best on all the datasets. GraphQNTK is not an R-convolution graph kernel either. It introduces some advanced techniques such as quantum computing and multi-head quantum attention mechanism to graph neural tangent kernels. It is also outperformed by MWSP on all the datasets.

Compared with the R-convolution graph kernels MLG, SP, WL, GWL, and GAWL, which do not solve both of the two challenges, MWSP achieves the best results on all the datasets. Like our method MWSP, both the R-convolution graph kernels WWL and FWL embed each graph into the vector space and consider each graph as a set of node embeddings. The similarity between two graphs is derived by computing the similarity between their node embeddings using the Wasserstein distance. MWSP outperforms both of them on all the datasets, which proves that the multi-scale shortest-path node feature map used in MWSP captures more information than the subtree pattern used in WWL and FWL.

\subsubsection{Ablation Study}
In this section, we compare MWSP to its two variants, i.e., WSP and MWSP-GFM. WSP is a kind of shortest-path graph kernel that is derived by not using the multiple different scales of the graph structure in MWSP. MWSP-GFM uses the multi-scale shortest-path graph feature maps rather than the multi-scale shortest-path node feature maps as used in MWSP. The classification accuracy of each method is given in Table~\ref{tab:ablation}. Compared with WSP, MWSP achieves the best results on 7 out of 14 datasets. On the remaining 7 datasets, MWSP and WSP have the same performance. Social networks usually have multiple different scales of the graph structure. MWSP is better than WSP on the IMDB-B dataset. For the REDDIT-B dataset, because MWSP runs out of memory under the parameter setting $k=1$, we do not conduct experiments with $k\geq 1$. Thus, MWSP and WSP have the same classification accuracy on the REDDIT-B dataset. This also happens on the DD dataset. For the other five datasets, because they do not have obvious multiple different scales of the graph structure, MWSP and WSP perform equally well. This proves that using the multiple different scales of the graph structure can improve performance and adaptiveness. Compared with MWSP-GFM, MWSP achieves the best results on 13 out of 14 datasets, which proves that considering the distributions of the shortest paths around each node is better. On the MUTAG dataset, MWSP-GFM has the same average classification accuracy as MWSP but a higher standard deviation than MWSP. Using multi-scale shortest-path node feature maps can lead to smoother classification accuracy across the 10 folds than using multi-scale shortest-path graph feature maps. We can see that MWSP is better than MWSP-GFM on 7 datasets in terms of the standard deviation of the classification accuracy over the 10 folds. MWSP and MWSP-GFM have the same standard deviation on two datasets NCI109 and PROTEINS. Compared with SP, it is evident that all of MWSP, WSP, and MWSP-GFM perform better, which proves that our strategies to improve the performance of SP are effective.

\begin{table}[!htb]
	\centering
	\caption{Comparison of classification accuracy (mean $\pm$ standard deviation) of MWSP to WSP (MWSP without using the multiple different scales of the graph structure to augment nodes), MWSP-GFM (MWSP using multi-scale shortest-path graph feature maps instead of multi-scale shortest-path node feature maps), and SP on the benchmark datasets.}
	\label{tab:ablation}
	\begin{tabular}{l|l|l|l|l}
		\toprule
		Dataset             &MWSP                                      &WSP                    &MWSP-GFM                 &SP   \\ \hline
		MUTAG             &\textbf{89.9$\pm$5.0}     &\textbf{89.9$\pm$5.0}   &\textbf{89.9$\pm$5.6}   &84.4$\pm$1.7    \\
		DHFR\_MD       &\textbf{73.5$\pm$5.2}      &\textbf{73.5$\pm$5.2}   &68.7$\pm$1.5   &68.0$\pm$0.4      \\
		NCI1                   &\textbf{86.3$\pm$1.1}     &\textbf{86.3$\pm$1.1}   &86.1$\pm$1.3    &73.1$\pm$0.3     \\
		NCI109              &\textbf{86.4$\pm$1.4}      &86.1$\pm$1.7   &85.9$\pm$1.4    &72.8$\pm$0.3      \\
		PTC\_MM         &\textbf{71.1$\pm$6.5}        &68.7$\pm$7.2   &70.2$\pm$5.4  &62.2$\pm$2.2        \\
		PTC\_MR          &\textbf{67.1$\pm$4.0}     &63.1$\pm$5.4   &65.1$\pm$5.3     &59.9$\pm$2.0     \\
		PTC\_FM           &\textbf{67.7$\pm$4.1}     &64.8$\pm$5.4   &65.3$\pm$3.1    &61.4$\pm$1.7     \\
		PTC\_FR            &\textbf{70.1$\pm$5.3}      &69.8$\pm$4.4   &69.3$\pm$4.3    &66.9$\pm$1.5       \\
		ENZYMES         &\textbf{66.7$\pm$4.7}      &\textbf{66.7$\pm$4.7}   &63.0$\pm$6.2    &41.1$\pm$0.8      \\
		PROTEINS        &\textbf{76.6$\pm$3.4}       &\textbf{76.6$\pm$3.4}   &76.3$\pm$3.4    &75.8$\pm$0.6       \\
		DD                      &\textbf{80.3$\pm$3.3}      &\textbf{80.3$\pm$3.3}   &80.1$\pm$3.0 &79.3$\pm$0.3  \\
		IMDB-B   &\textbf{75.2$\pm$2.7}      &74.4$\pm$4.8   &74.8$\pm$3.7 &72.2$\pm$0.8  \\
		REDDIT-B   &\textbf{90.8$\pm$1.2}      &\textbf{90.8$\pm$1.2}   &86.6$\pm$2.6 &84.6$\pm$0.2  \\
		Cuneiform         &\textbf{79.8$\pm$0.7}        &78.8$\pm$1.3   &79.2$\pm$1.4    &74.4$\pm$0.9       \\
		\bottomrule
	\end{tabular}
\end{table}

\subsubsection{Parameter Sensitivity}\label{ps}

\begin{figure}[!htb]
	\centering
	\subfigure[DHFR\_MD]{\includegraphics [width=0.24\textwidth]{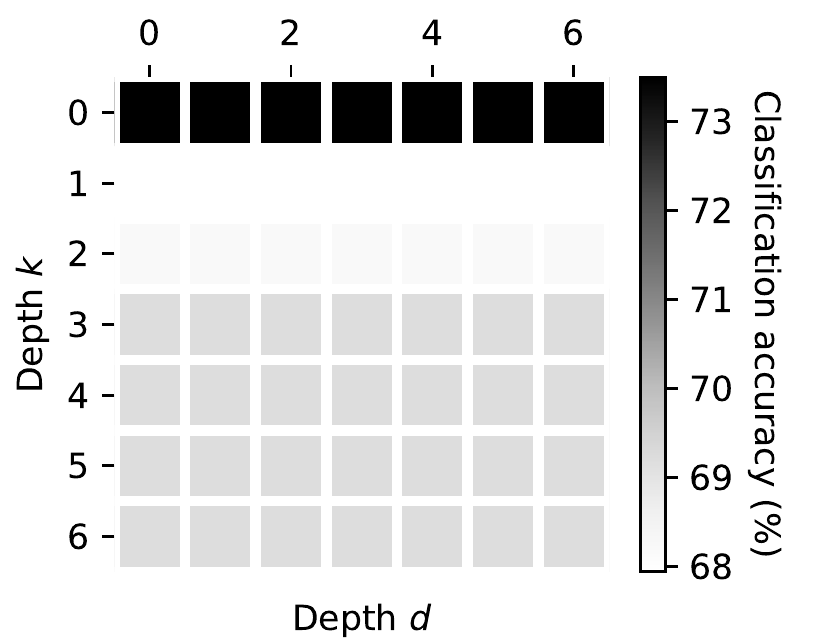}}
	\centering
	\subfigure[PTC\_FM]{\includegraphics [width=0.24\textwidth]{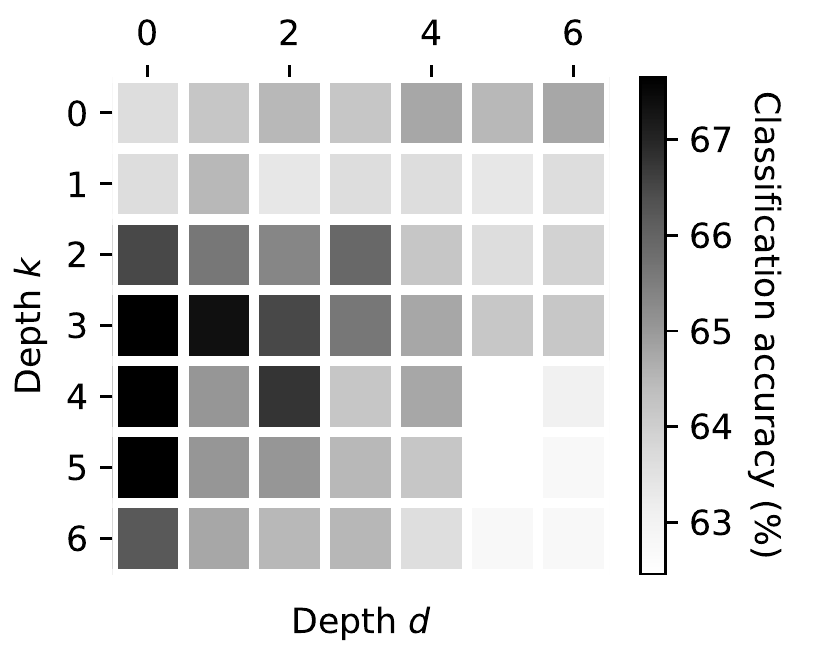}}
	
	\centering
	\subfigure[IMDB-B]{\includegraphics [width=0.24\textwidth]{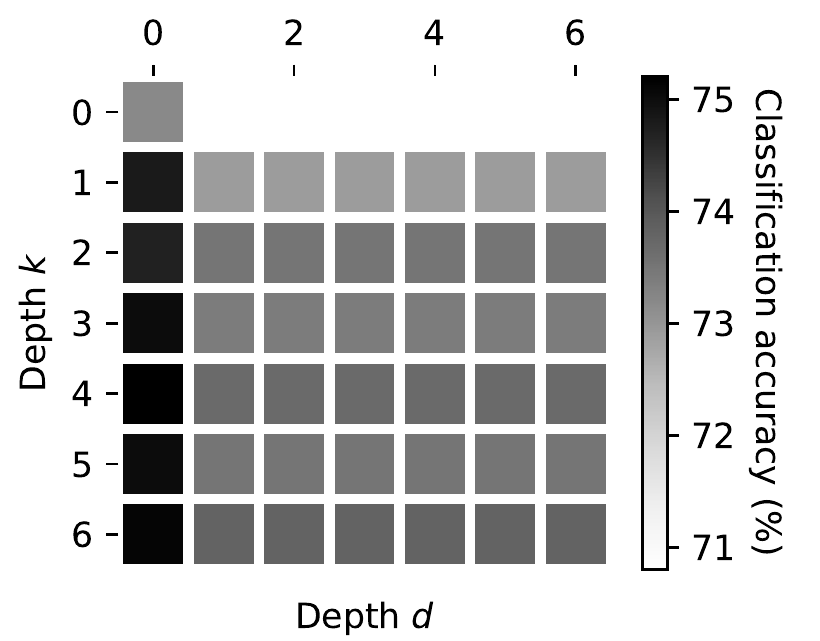}}
	\centering
	\subfigure[Cuneiform]{\includegraphics [width=0.24\textwidth]{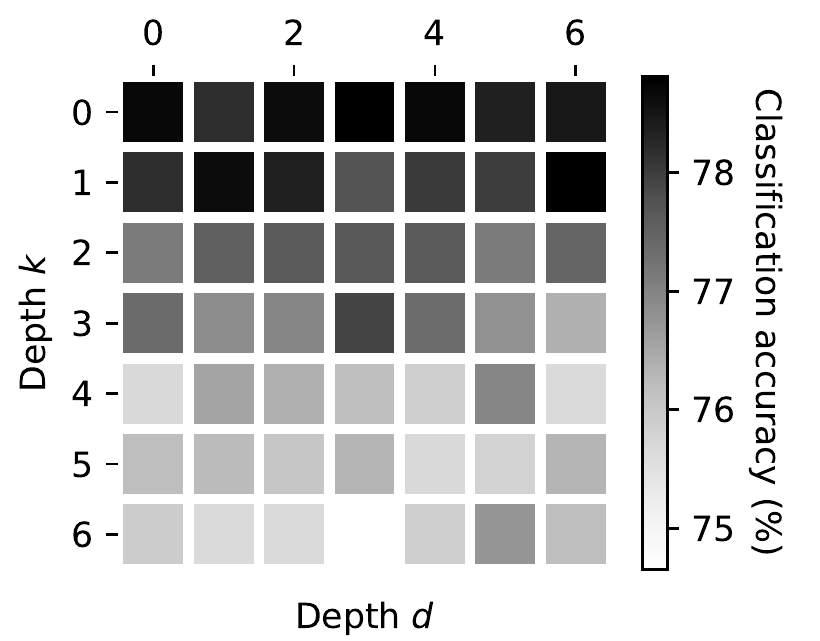}}
	\caption{Parameter sensitivity of MWSP. $d$ represents the depth of the truncated BFS tree rooted at each node, which is used for the extraction of shortest paths whose maximal length is $d$. $k$ represents the depth of the truncated BFS tree rooted at each node, which is used for capturing multiple different scales of the graph structure around each node (for augmenting graph nodes). Each entry in the heatmap is the mean classification accuracy over the 10 folds.}
	\label{fig:param_sensitivity}
\end{figure}

To analyze parameter sensitivity with regard to the two parameters $d$ and $k$ of MWSP, we compute average classification accuracy over the 10 folds for all the combinations of $d$ and $k$, where $d, k \in \{0, 1, 2, \ldots, 6\}$, on datasets DHFR\_MD, PTC\_FM, IMDB-B, and Cuneiform. DHFR\_MD is a chemical compound dataset. PTC\_FM is a molecular compound dataset. IMDB-B is a social network. Cuneiform is a Hittite cuneiform signs dataset. We select these four datasets to represent their categories. Figure~\ref{fig:param_sensitivity} shows the heatmaps of the results. On datasets DHFR\_MD and IMDB-B, the classification accuracy remains stable when the values of $k$ and $d$ exceed a threshold. To explain this phenomenon, we check the multi-scale shortest-path feature map of each node in datasets DHFR\_MD and IMDB-B and find that its dimension does not change when the values of $k$ and $d$ exceed a threshold. We can see from Table~\ref{tab:statistics_dataset} that both the mean shortest-path length and the maximal shortest-path length of DHFR\_MD equal one. A BFS tree of depth one rooted at each node can capture all the shortest-path patterns in DHFR\_MD. As for the IMDB-B dataset, because the maximal shortest-path length is two, a BFS tree of depth two rooted at each node can capture all the shortest-path patterns in IMDB-B. The best parameters for these four datasets are different owing to their different statistics as shown in Table~\ref{tab:statistics_dataset}. To sum up, the best value of $k$ or $d$ is around the mean shortest-path length of these four datasets.

\begin{table*}[!htb]
	\centering
	\caption{Comparison of CPU running time for computing the kernel matrix of MWSP to other graph kernels on the benchmark datasets. N/A means the result is not available because the current code of FWL does not support that specific dataset.}
	\label{tab:time}
	\begin{tabular}{l|l|l|l|l|l|l|l|l|l|l}
		\toprule
		Dataset             &MWSP     &MLG                  &RetGK                 &SP            &WL             &FWL       &WWL              &GWL      &GAWL   &GraphQNTK\\ \hline
		MUTAG             &3.15''       &1'18''                 &1.34''                            &0.23''       &0.20''         &9.63''     &5.51''              &0.05''     &2.02''   &10.83''\\
		DHFR\_MD       &19.16''      &6'16''                &3.74''                            &1.11''         &0.94''         &34.06''   &29.98''          &0.25''      &3.44''  &3'5''\\
		NCI1                   &47'31''     &59'19''              &12'5''                            &16.26''     &17.21''         &56'41''   &1h 8'52''        &9.45''      &10'14'' &1h 59'51''\\
		NCI109              &46'56''     &1h 3'                 &11'21''                            &16.09''     &16.77''         &1h 52''   &1h 12'21''        &9.65''      &10'15''&1h 55'23''\\
		PTC\_MM         &8.86''        &1'57''                &4.20''                            &0.33''       &0.34''          &18.44''   &14.86''            &0.07''     &2.16''   &30.00''\\
		PTC\_MR          &9.59''        &1'54''                &4.60''                            &0.35''       &0.36''           &19.28''   &16.35''           &0.07''     &2.19''   &31.50''\\
		PTC\_FM           &9.94''       &1'54''                &4.66''                            &0.33''        &0.39''           &21.19''   &15.95''           &0.07''     &2.26''   &32.35''\\
		PTC\_FR            &9.96''       &1'50''                &4.73''                            &0.35''        &0.39''           &20.52''   &26.57''           &0.13''     &2.32''   &36.04''\\
		ENZYMES         &58.92''      &9'29''               &17.42''                            &2.54''         &3.95''          &1'16''       &1'35''             &0.44''    &16.22''  &3'33''\\
		PROTEINS        &5'                &31'12''              &1'6''                            &14.99''        &17.62''        &4'17''      &6'52''            &0.68''    &54.91''  &14'9''\\
		DD                      &9h 31'10''  &13h 26'50''     &12'43''                            &1h 15'35''   &17'40''         &31'29''   &4h 52'55''    &10.68''   &15'47''  &19h 19'23''\\
		IMDB-B            &8'51''            &25'15''            &0.33''                            &2.42''          &3.61''           &2'46''     &3'12''             &0.35''    &12.08''  &9'56''\\
		REDDIT-B        &29h 2'26''    &$>$72h          &2'29''                            &$>$48h        &17'35''         &43'35''   &44h 17'45''   &8.12''    &12'42''  &$>$72h\\
		Cuneiform         &6.61''           &3'2''                &1.65''                            &0.45''         &0.47''           &N/A          &12.29''             &0.08''   &2.32''   &15.97''\\
		\bottomrule
	\end{tabular}
\end{table*}

\subsubsection{Running Time}
In this section, we report the running time needed to compute the kernel matrices of MWSP and other graph kernels on the benchmark datasets in Table~\ref{tab:time}. Since the time complexity of MWSP is linear with respect to the parameters $k$ and $d$ (please see Section~\ref{complex}), the running time of MWSP will increase linearly with regard to $k$ or $d$. For convenience, the values of $k$ and $d$ are set to zero and one, respectively. Note that GWL is written in Java and runs the fastest. All the other methods are written in Python. Since the four R-convolution graph kernels SP, WL, GWL, and GAWL do not consider the distributions of substructures in the computation of graph kernels, they also run very fast on most datasets. RetGK uses a random Fourier feature map~\cite{rahimi2007random} to approximate the kernel matrix computation with promising running time. MLG develops a randomized projection method, which is similar to the Nystr\"{o}m method, to approximate the kernel matrix computation. However, MLG runs slower than WWL. Both MWSP and WWL use the Wasserstein distance to compute the graph similarity and they have a similar running time. FWL runs faster on large datasets compared with MWSP and WWL. Although GraphQNTK adopts quantum parallelism to accelerate the kernel matrix computation, it has a worse running time compared with WWL and MWSP, especially on large datasets. Our method MWSP has a similar time complexity to WWL. One possible solution to reduce the running time of MWSP is to use the Sinkhorn method~\cite{altschuler2017near,cuturi2013sinkhorn}.

\section{Conclusion}\label{conclusion}
In this paper, we have proposed a novel muti-scale graph kernel called MWSP to deal with the two challenges of the R-convolution graph kernels, i.e., they cannot compare graphs at multiple different scales and do not consider the distributions of substructures. To mitigate the first challenge, we propose the shortest-path graph/node feature map, whose element represents the frequency of each shortest path in a graph or around a node. Since the shortest-path graph/node feature map cannot compare graphs at multiple different scales, we incorporate into it the multiple different scales of the graph structure, which are captured by the truncated BFS trees of different depths rooted at each node in a graph. To mitigate the second challenge, we adopt the Wasserstein distance to compute the similarity between the multi-scale shortest-path node feature maps of two graphs. Experimental results show that MWSP outperforms the state-of-the-art graph kernels on most datasets. Since the Wasserstein distance has a high time complexity, we would like to develop an efficient distance measure to compute the similarity between the node feature maps of two graphs in the future.

%\section*{Acknowledgment}
%This work was supported partially by the National Natural Science Foundation of China (grant \# 61796184), the National Key Research and Development Program of China (grant \# 2020AAA0108100), and the Fundamental Research Funds for the Central Universities of China.

\bibliographystyle{IEEEtran}
\bibliography{reference}

\end{document}